\documentclass[fleqn,11pt]{wlscirep}
\usepackage[utf8]{inputenc}
\usepackage[T1]{fontenc}
\usepackage{hyperref}
\usepackage{color,soul}

\title{Learning nonlinear integral operators via Recurrent Neural Networks and its application in solving Integro-differential Equations}

\author[1] {Hardeep Bassi}
\author[2]{Yuanran Zhu}
\author[2]{Senwei Liang}
\author[2]{Jia Yin}
\author[3]{Cian C. Reeves}
\author[4, 5]{Vojtěch Vlček}
\author[2]{Chao Yang}
\affil[]{Department of Applied Mathematics, University of California, Merced,  Merced, USA, 95340}
\affil[2]{ Applied Mathematics and Computational Research Division, Lawerence Berkeley National Laboratory, Berkeley, USA, 94720}
\affil[3]{Department of Physics, University of California, Santa Barbara,  Santa Barbara, USA, 93117}
\affil[4] {Department of Chemistry and Biochemistry, University of California, Santa Barbara,  Santa Barbara, USA, 93117}
\affil[5] {Department of Materials, University of California, Santa Barbara,  Santa Barbara, USA, 93117}

\begin{abstract}
In this paper, we propose using LSTM-RNNs (Long Short-Term Memory-Recurrent Neural Networks) to learn and represent nonlinear integral operators that appear in nonlinear integro-differential equations (IDEs). The LSTM-RNN representation of the nonlinear integral operator allows us to turn a system of nonlinear integro-differential equations into a system of ordinary differential equations for which many efficient solvers are available. Furthermore, because the use of LSTM-RNN representation of the nonlinear integral operator in an IDE eliminates the need to perform a numerical integration in each numerical time evolution step,  the overall temporal cost of the LSTM-RNN-based IDE solver can be reduced to $O(n_T)$ from $O(n_T^2)$ if a $n_T$-step trajectory is to be computed.  We illustrate the efficiency and robustness of this LSTM-RNN-based numerical IDE solver with a model problem. Additionally, we highlight the generalizability of the learned integral operator by applying it to IDEs driven by different external forces. As a practical application, we show how this methodology can effectively solve the Dyson's equation for quantum many-body systems.
\end{abstract}
\begin{document}

\flushbottom
\maketitle
%
%
\thispagestyle{empty}

\section{Introduction}
Integro-differential equations (IDEs) arise in many scientific applications ranging from nonequilibrium quantum dynamics\cite{stefanucci2013nonequilibrium,cohen2011memory,reeves2023unimportance,Reeves_2023}, the dynamics of non-Markovian colloidal particles \cite{zhu2022_EMZ,zhu2018estimation,zhu2020generalized,zwanzig2001nonequilibrium} the modeling of dispersive waves \cite{whitham1967variational,debnath2005nonlinear} and electronic circuits\cite{bohner2022qualitative}. One particularly important example is the application of the Kadanoff-Baym equation \cite{kadanoff2018quantum,stefanucci2013nonequilibrium,reeves2023unimportance} for the time evolution of quantum correlators. This equation describes the propagation in time of non-equilibrium Green's functions, and finding efficient methods of solving this equation is of extreme importance in the study of driven quantum systems. A general type of IDEs can be written as:
\begin{equation}\label{eqn:IDE}
\frac{d}{dt}G(t) = F(G(t),t) +\int_0^t K(G(t-s),s)G(s)ds
\end{equation}
where both $F$ and the integral kernel $K$ are functions of and $t$ and $G(t)$. An IDE is computationally challenging to solve due to the presence of the integral term in \eqref{eqn:IDE}.  A numerical time evolution scheme typically requires performing a numerical integration of the integral term at each time step.  If $n_T$ time steps are taken to evolve the numerical solution of \eqref{eqn:IDE} to time $T$, the overall computational complexity in time is proportional to at least $n_T^2$, which can be high for a large $n_T$.

Several techniques have been recently developed to reduce the temporal complexity of solving \eqref{eqn:IDE}.  One technique is based on constructing and updating a compact representation of $G(t)$ and $K(G(t-s),s)$~\cite{kaye2023fast}. Another technique uses snapshots of the solution to \eqref{eqn:IDE} within a small time window to construct a reduced order model that can be used to extrapolate the long-time dynamics of $G(t)$~\cite{DMDdiag,Reeves_2023}. 
In this paper, we present yet another approach to reducing the computational cost of solving \eqref{eqn:IDE}. The basic idea of this approach is to turn an IDE such as \eqref{eqn:IDE} into an ordinary differential equation (ODE) of the form
\begin{equation}\label{eqn:ODE} 
\frac{d}{dt}G(t) = F(G(t),t) + I(G(t),t),
\end{equation}
where $I(G(t),t)$ is a functional of $G(t)$ and $t$ that can be evaluated with a constant cost, i.e., without performing numerical integration. A standard ODE scheme can then be applied to solve \eqref{eqn:ODE} with a temporal complexity of $O(n_T)$.

In principle, the mapping from $G(t)$ to $I(G(t),t)$,  which is implicitly defined by the integral $\int_0^t K(G(t-s),s)G(s)ds$ always exists, although its (memoryless) analytical form is generally unknown. In this work, we seek to represent and learn such a mapping by training a recurrent neural network (RNN) using snapshots of the numerical solution of \eqref{eqn:IDE} within a small time window. Hence our approach falls into  the category of operator learning methods, which encompasses recently developed machine-learning methodologies such as  DeepONet\cite{lu2021learning} and the Fourier neural operator\cite{li2020fourier,kovachki2023neural}. In an operator learning method, we view the solution to a given ODE or partial differential equation (PDE) as the output of a neural network parameterized operator that maps between function spaces. For instance, solving an initial value problem for a given PDE can be reformulated as searching for a neural network parameterized operator, denoted as $S: u(x,0) \rightarrow u(x,t)$. Carefully designed neural networks, such as those employed in DeepONet and the Fourier neural operator, can approximate this operator $S$. By applying the learned operator $S$ to the initial condition $u(x,0)$ we can readily obtain the solution to the PDE. One notable advantage of the operator learning approach lies in the generalizability of the trained neural network model. Once the approximation to the operator is obtained, it can be applied to other initial conditions or inputs to the model.

In this work, we adopt an operator-learning perspective and choose to use a long-short term memory (LSTM)-based RNN~\cite{hochreiter1997long} to learn the mapping between $G(t)$ and $I(G(t),t)$. Instead of a simple feed-forward neural network (FFNN), we utilize RNNs because they can better preserve the causality of both $G(t)$ and $I(G(t),t)$. Furthermore, instead of learning the operator that yields the solution to the IDE \eqref{eqn:IDE} directly, we choose to learn the mapping between $G(t)$ and $I(G(t),t)$ for the following reasons. First of all, although the solution of \eqref{eqn:IDE} depends on both $F(G(t),t)$ (referred to as the {\em streaming term}) and $I(G(t),t)$ (referred to as the {\em memory integral} or {\em collision integral}), the cost of evaluating the memory term typically far exceeds that of the streaming term. Secondly, when the streaming term $F(G(t),t)$ is time-dependent and creates atypical dynamics outside of the training window (see e.g. Figure \ref{fig:multi_traj_toy_model_Ct_real}), it is difficult to directly learn a solution operator using short-time data. Consequently, the extrapolated prediction of the solution of \eqref{eqn:IDE} for large $t$ can be poor. On the other hand, in many physically relevant scenarios, the integral kernel in \eqref{eqn:IDE} is well behaved. As a result, we expect the mapping between $G(t)$ and $I(G(t),t)$, which is independent of the solution $G(t)$ itself, can be learned more easily. Furthermore, in addition to increasing the time window and training the RNN with different initial conditions, we can augment the training data by considering solutions of \eqref{eqn:IDE} with different streaming terms within a small time window.  We will refer to this type of training as multi-trajectory training in section~\ref{sec:numerical_results}. This type of training is important for learning an operator that maps from one function space to another. Once the integral operator $I(t)$ is well approximated by a properly trained RNN, the solution of the IDE can be obtained using any standard ODE solver with $I(t)$ evaluated by the RNN. Furthermore, transferability of the learning algorithm is clearly achievable. 
Once the integral operator with a fixed integral kernel $K$ is learned via an LSTM-based RNN, it can be used to solve the IDEs~\eqref{eqn:IDE} associated with different streaming forces $F(G(t),t)$ by using a standard ODE solver. 

The approach presented in this work is similar in spirit to other developments in using machine learning (ML) techniques and neural networks (NN) to solve forward and inverse problems defined by ODEs and PDEs. The most prominent example 
is the physics-informed neural network (PINN)~\cite{chen2021learning,zhang2019quantifying} and its generalizations. In the PINN approach, an NN serves as a solver that takes the spatial-temporal coordinate $x,t$ as the input and outputs the approximate solutions to the differential equation. The whole network is trained using the loss function that is defined in terms of the underlying differential equation. More recent members within the PINN family include sparse physics-informed neural network (SPINN)~\cite{RAMABATHIRAN2021110600} and parareal physics-informed neural network (PPINN)~\cite{meng2020ppinn}.
Another representative approach is the neural ordinary differential equation (NeuralODE) ~\cite{chenNeuralOrdinaryDifferential2019} and it is particularly useful for solving the inverse problem of differential equations. In this approach, an NN is used to approximate the derivatives of the state variables, i.e. the "right-hand side" of a differential equation. After the recovery of the derivatives, solutions to the differential equations can be obtained using a standard numerical solver for ODEs and PDEs. Recent developments in the NeuralODE family include NeuralSDE~\cite{liuNeuralSDEStabilizing2019,li2020scalable}, Neural Jump SDE~\cite{jiaNeuralJumpStochastic2020}, NeuralSPDE~\cite{salvi2021neural}, Neural Operators~\cite{kovachki2021neural}, infinitely deep Bayesian neural networks~\cite{xuInfinitelyDeepBayesian2021}, \textit{etc}.

What distinguishes the approach presented in this work from other existing neural-network-based differential equation solver is that we do not use the RNN to solve the IDEs directly. Instead, we use it to approximate the time dependent nonlinear integral operator that is costly to evaluate. We combine the RNN based operator learning with a standard ODE solver to obtain numerical solutions to the IDEs.

This paper is organized as follows. In Section \ref{sec:methods}, we first introduce the basic architecture of RNN and the LSTM model, then we customize a specific RNN model designed to approximate the integral functional $I(t)$ and introduce two training methods for the RNN model. In Section \ref{sec:numerical_results}, we use a nonlinear, complex-valued IDE for $2\times 2$ matrix $G(t)$ as an example to demonstrate the effectiveness of using RNN to learn the integral operator and how this can be used to extrapolate the dynamics of the IDE. Importantly, we showcase the remarkable ability of using the same integral operator to solve IDEs driven by different streaming terms, thus affirming its generalizability. In Section \ref{sec:dyson_eqn}, we will consider a specific physical application of the introduced methodology. Here, we underscore the practical utility of the RNN model in solving the Dyson's equation, offering further insights into its applicability. The primary findings and conclusions of this paper are succinctly summarized in Section \ref{sec:conclusion}.






%
%
%
\section{Methods}\label{sec:methods}
\subsection{Recurrent Neural Network}

Recurrent neural networks (RNN) and their various adaptations have become prevalent tools to analyze and forecast the patterns in time series data~\cite{zhang1998time,karim2017lstm,hu2020time} across a wide range of domains and scenarios~\cite{cho2014learning, can2018multilingual}. In this section, we will introduce the basic RNNs model, as well as the LSTM model.


\begin{figure}[t]
    \centering
    \includegraphics[width=0.75\linewidth]{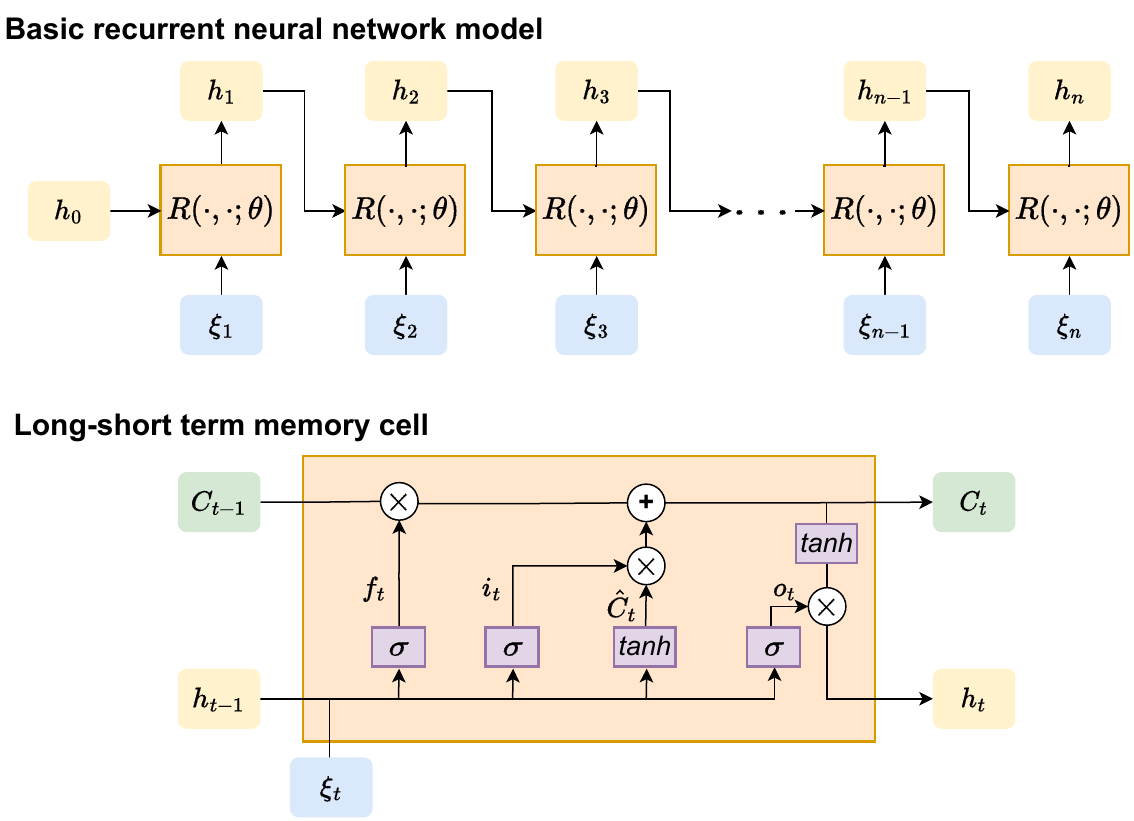}
    \caption{\textbf{(Top) The workflow of a basic RNN model. } The model processes the inputs $\{\xi_1, \xi_2, \cdots , \xi_n\}$ one by one in sequence and produces a corresponding sequence of hidden states $\{h_1, h_2, \cdots , h_n\}$. The same set of model parameters, $\theta$, is employed in the RNN cell (i.e., $R(\cdot,\cdot;\theta)$) to calculate each $h_i$, $i=1,\cdots,n$. \textbf{(Bottom) An LSTM cell}. The update of the cell state $C_t$ and the hidden state $h_t$ relies on the current time input  $\xi_{t}$ and the preceding states, $h_{t-1}$ and $C_{t-1}$, through several information gates.}
    \label{fig:basic_RNN}
\end{figure}

\paragraph{Basic RNN model} The workflow of the RNN model is illustrated in Figure \ref{fig:basic_RNN}~(Top), which comprises several key components: including inputs, hidden states, and an RNN cell. The hidden states are calculated recursively within the RNN cell, utilizing sequential inputs. These hidden states play an important role in capturing temporal dependencies and facilitating the propagation of information over time within sequential data. Furthermore, these hidden states can be employed to model time-dependent quantities of interest, such as the memory integral $I(G(t),t)$, which is the main focus of our paper.

Specifically, consider a time series dataset $\boldsymbol{\Xi} = \{\xi_{0}, \xi_{1}, \xi_{2}, \ldots , \xi_{N}\}$, where $\xi_{i}$ represents the $i^{th}$ time index within our time series. The RNN cell, represented as a function $R(\cdot,\cdot;\theta)$, takes the preceding hidden state and the current time data as input and produces a new hidden state as output. Namely, $h_{t} = R(\xi_{t}, h_{t-1};\theta), t=1,\cdots,N$. Here, $h_0$ is initialized as a zero vector and $\theta$ represents the set of shared and trainable parameters. Mathematically, the hidden states can be written as a composition of the RNN cells given by:
\begin{equation*}
    h_{n} = R(\xi_{n}, h_{n-1};\theta) = R(\xi_{n}, R(\xi_{n-1}, h_{n-2};\theta);\theta) = R(R( ... R(\xi_{2}, R{(\xi_{1}, h_0;\theta);\theta), ... ;\theta);\theta)}.
\end{equation*}
The sharing of $\theta$ enables the RNN to learn general patterns across time and incorporate a memory effect into the model. This capability also allows RNN models to handle input sequences of varying lengths.


\paragraph{Long-short term memory (LSTM)} The LSTM model is a type of RNN that distinguishes itself by employing different gates within its LSTM cell to regulate the flow of information, as shown in Figure \ref{fig:basic_RNN}~(Bottom). Compared with the standard RNN models, LSTM models exhibit better performance in capturing long-range dependencies and mitigating the vanishing gradient issue~\cite{hochreiter1997long}. Their gated structure allows them to effectively preserve information across longer sequences. As a result, LSTM models have gained widespread applications in dynamics modeling tasks~\cite{zhu2023learning, harlim2021machine}. In our specific context, we will employ an LSTM model to model the memory integral of the IDE.

The LSTM cell features two distinct states: the cell state, denoted as $C_t$, and the hidden state, represented as $h_t$. The cell state serves as a repository for long-term memory, capable of retaining and propagating information throughout different time steps. On the other hand, the hidden state captures current information, serving as the foundation for predictions made at each time step. To regulate the flow of state information, three pivotal gates are employed: the input gate, the forget gate, and the output gate. The forget gate is responsible for determining which information from the previous cell state should be retained in the current cell state, while the input gate governs the incorporation of new information into the current cell state. Concurrently, the output gate controls the outward transmission of information from the cell state.

Let $i_t$, $f_t$, and $o_t$ represent the input, forget, and output gates at time $t$. Each of these gates employs a nonlinear activation function, such as $\sigma(x) = \frac{1}{1+e^{-x}}$, in conjunction with learnable parameters to govern the selection of information to be incorporated into the state updates. This can be expressed as follows:
\begin{equation*}
i_t = \sigma(\xi_{t}, h_{t-1};\theta_{input}), \quad f_t = \sigma(\xi_t, h_{t-1};\theta_{forget}), \quad o_t = \sigma(\xi_t, h_{t-1};\theta_{output}),
\end{equation*}
Here, $\theta_{input}$, $\theta_{forget}$, and $\theta_{output}$ are parameter sets within total collection of shareable parameters $\theta$.

Using these gates, the current cell state is updated according to:
\begin{equation*}
    C_t = f_t \cdot C_{t-1} + i_t \cdot \hat{C}_t,
\end{equation*}
 which combines both long-term memory (e.g., $C_{t-1}$) and new information (e.g., $\xi_{t}$ used in the computation of $\hat{C}_t= tanh(\xi_t, h_{t-1}; \theta_{c})$). Here, $\theta_{c} \in \theta$ is the parameter set. Finally, we compute the current hidden state $h_t$ as:
\begin{equation*}
    h_t = o_t \cdot \tanh(C_t).
\end{equation*}
The hidden state $h_t$ can subsequently undergo further trainable transformations, such as a linear transformation, to predict the specific quantity of interest.

\begin{figure}[t]
    \centering
     \includegraphics[width=0.8\linewidth]{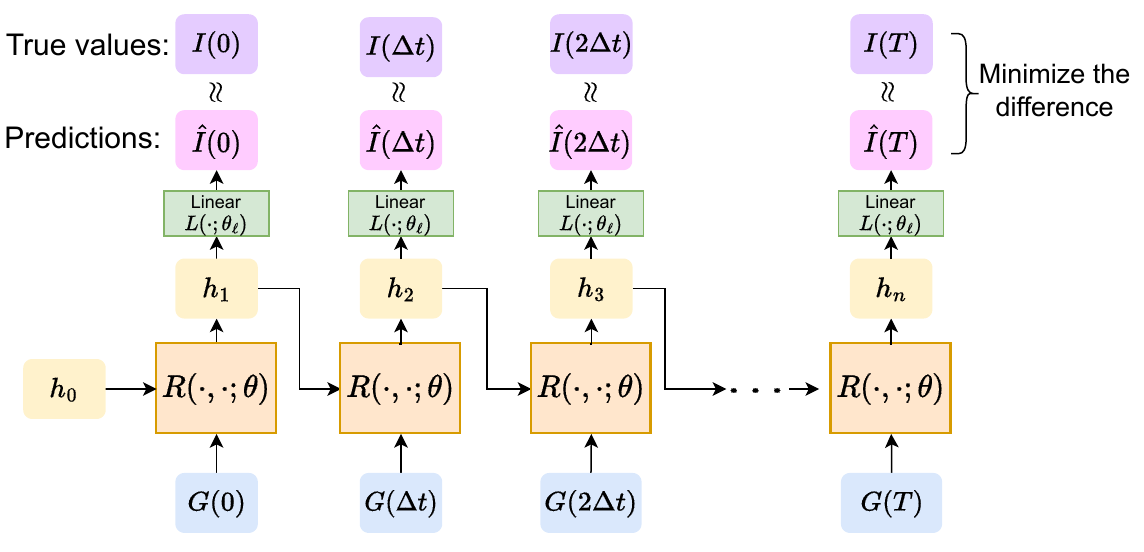}
    \caption{Learning and predicting diagram of the RNN model. In the training phase, the input sequence $\{G(0), G(\Delta t), G(2\Delta t), ... , G(T)\}$ is fed into the model to generate predictions $\{\hat{I}(0), \hat{I}(\Delta t), \hat{I}(2\Delta t), ... , \hat{I}(T)\}$, using the parameters $\theta$. From here, we can use the final output $\hat{I}(T)$ to produce numerically extrapolated dynamics using the procedure outlined in \textit{Model Architecture} and illustrated in Figure \ref{fig:customized_rnn_extrapolation}. }
    \label{fig:customized_rnn}
\end{figure}


\begin{figure}[t]
    \centering
     \includegraphics[width=0.9\linewidth]{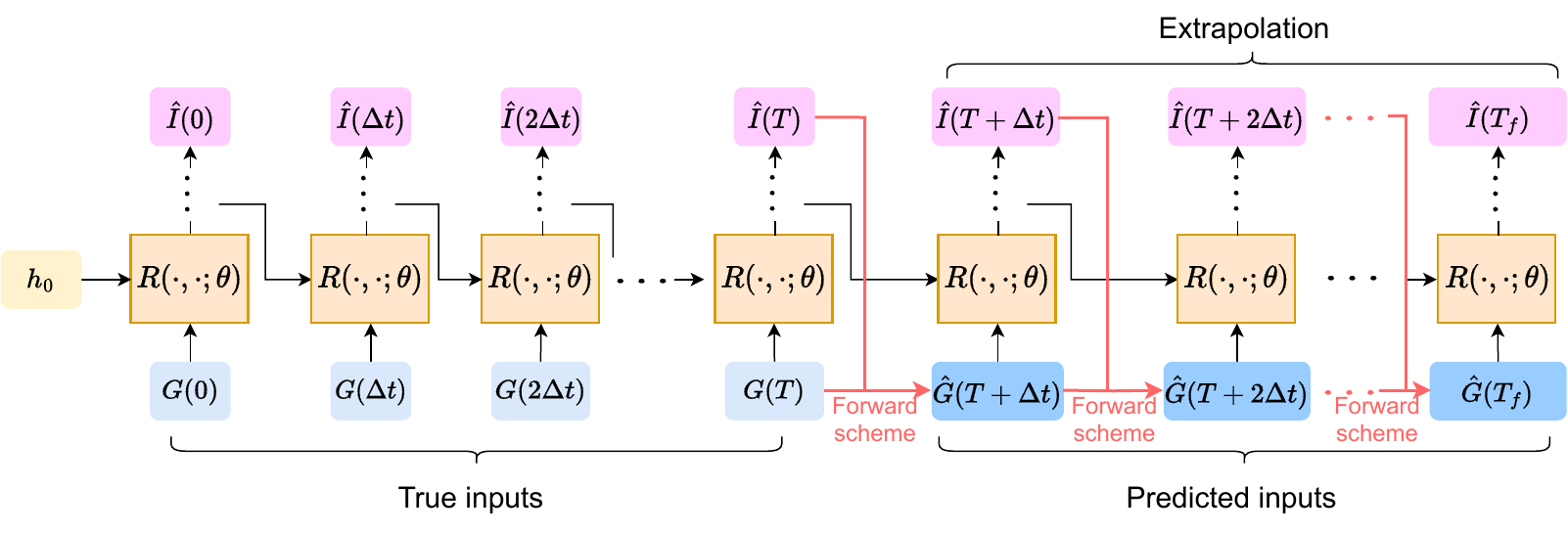}
    \caption{Extrapolation of dynamics using the RNN model. The training phase will output predictions $\{\hat{I}(i \Delta t)\}_{i=0}^{T}$. Using $\hat{I}(T)$, we can use a forward numerical method to obtain the numerically extrapolated $G(T + \Delta t)$. From here, we can recursively feed the numerically extrapolated $G(T + \Delta t)$  into the model to obtain the prediction for $\hat{I}(T + \Delta t)$. We repeat this until the final time, $T_f$. }
    \label{fig:customized_rnn_extrapolation}
\end{figure}


\begin{figure}[t]
    \centering
     \includegraphics[width=0.8\linewidth]{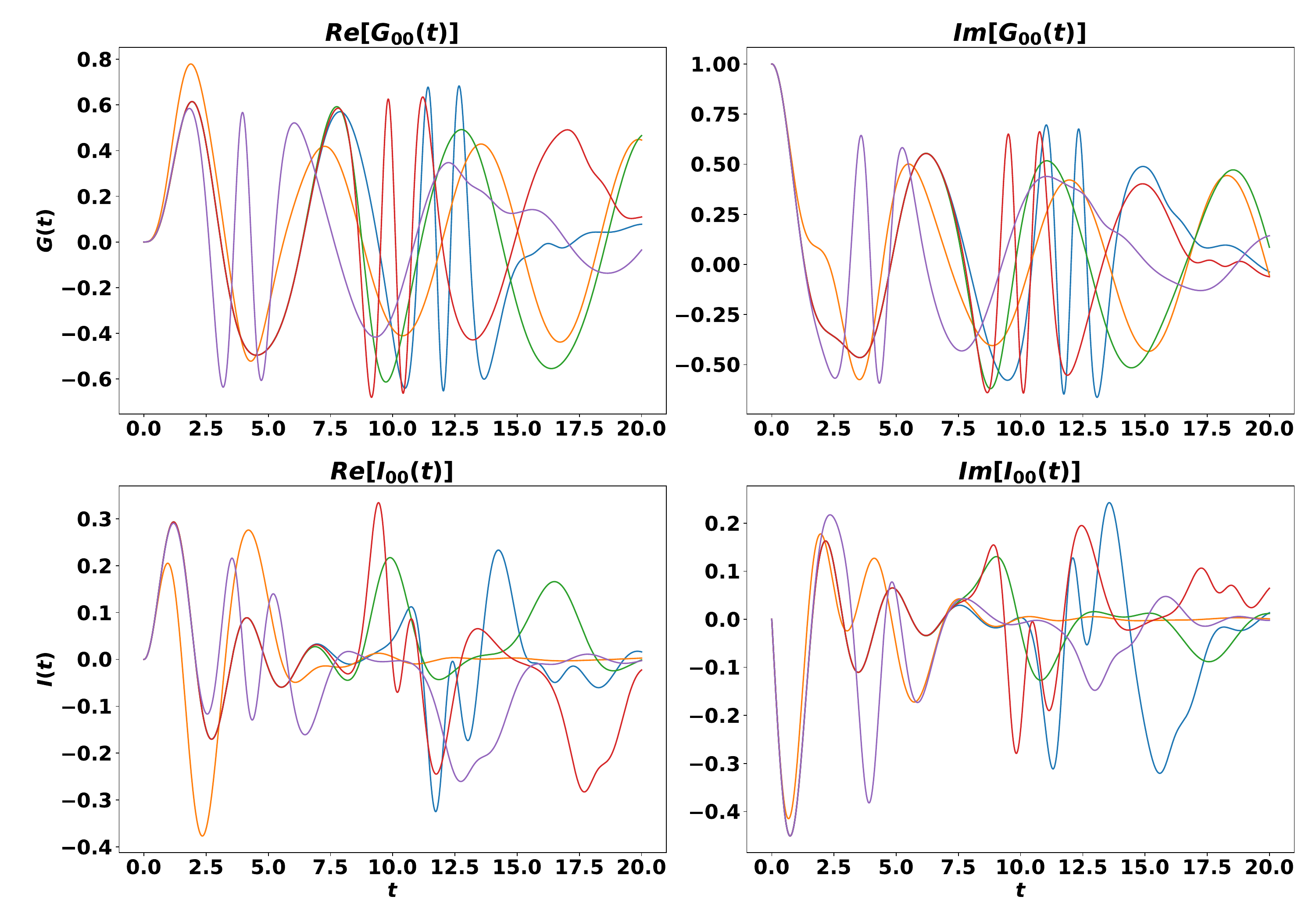}
    \caption{Sample trajectories of $G_{00}(t),I_{00}(t)$, selected from the database created by solving Eqn \eqref{eq:toymodel} with $\alpha_1, \alpha_2 \in [1,20]\times [1,20]$, $\sigma \in \{1,2,3,4,5\}$, and $\beta=1$. Subsequently, we use the same database to do multi-trajectory training of the RNN model.}
    \label{fig:toy_model_training_data}
\end{figure}
\subsection{RNN customized for learning time-dependent nonlinear integral operator}
\label{sec:customized_rnn}
Having introduced the basic architecture of RNN and the integro-differential equation we aim to solve, in this section, we customize a specific RNN model that enables us to efficiently learn the integral operator of an IDE within a short time window and use that to predict its long-time dynamics. As mentioned in the introduction, the RNN we seek should learn a map $I: G(t)\rightarrow I(t)$ for any given function $G(t)$. Since $I(t)$ depends on all the history values of $G(t)$, to capture this memory effect, we propose to use the LSTM cells as the basic modeling modules to build the RNN model. This leads to a {\em learning-predicting} diagram as illustrated in Figure \ref{fig:customized_rnn}.
\paragraph{Model architecture and dynamics extrapolation}
The RNN in Figure \ref{fig:customized_rnn} consists of an LSTM model and a linear transformation layer attached to it that maps the output of LSTM cells into the shape of the target time series. For our model, the NN takes the discretized $G(t)$ in a time grid $t=i\Delta t$ as the input and produces an approximated solution of the collision integral $\hat I(t)$ for $t=i\Delta t$ as the output. The RNN model is trained using a sufficiently accurate numerical solution to the IDE within a time window $t\in[0,T]$. After the training is done, in the extrapolation phase, we can use an efficient and accurate forward propagation scheme to generate long-time trajectories. As an example, if we employ the forward Euler scheme to solve the differential equation, then for IDE \eqref{eqn:ODE}, we have  
\begin{align*}
G(T+(i+1)\Delta t)=G(T+i\Delta t) +\Delta t[F(T+i\Delta t,G(T+i\Delta t))+\hat I(T+i\Delta t)]
,\qquad i\geq 0,
\end{align*}
where $\hat I(T+i\Delta t)$ is the output of the RNN generated recursively by feeding in the input $G(T+i\Delta t)$ as illustrated in Figure \ref{fig:customized_rnn_extrapolation}. The forward scheme for any multi-step method can be similarly derived. The specific model parameters, such as the hidden size of each layer,
will be discussed in Section \ref{sec:numerical_results} per IDE considered. 
\paragraph{Data preparation and Loss functions}
To generate the training data, we solve the IDE numerically using an Adams-Bashforth third-order method (AB3) and Simpson's rule to approximate the collision integral $I(t)$. This generates a time series data that is recorded in an interval of $[0,T]$, discretized by $\Delta t$, eventually forming a dataset consisting of $\{(G(i\Delta t), I( i\Delta t)\}_{i=0}^T$ pairs. For the complex-values IDEs that are considered in our applications, we will further split the real and imaginary parts of $G(t)$ and $I(t)$ when we generate the input sequence and compare $\hat I(t)$ with $I(t)$. This has to be done since most popular ML frameworks such as PyTorch\cite{NEURIPS2019_9015} can only do real-valued arithmetics when training the NN. The parameters of the network, denoted by $\theta$, are randomly initialized. The Adam optimizer\cite{KingBa15}, which is
a first-order method that uses gradient-based optimization, is chosen to adjust the parameter values to minimize the mean-squared error (MSE) function: 
\begin{equation} \label{eq:loss}
    f(I, \hat I;\theta) := \frac{1}{N}\sum_{i=0}^N(I(i \Delta t) - \hat I(i\Delta t))^2,
\end{equation} 
where $\hat I(i\Delta t)$ is the predicted collision integral generated by the RNN and $I(i\Delta t)$ is the ground truth. The parameters are learned by propagating the gradients of each hidden state’s inputs. 

\paragraph{Training method}
As we mentioned before, for all the numerical simulations considered in this paper, the RNN training is performed in a small time window ($T$ relatively small), and the trained model is used for long-time ($T_f$) extrapolation in which $T_f \gg T$, see Figure \ref{fig:customized_rnn_extrapolation}. We employ two training strategies to learn the integral operator $I(t)$:
\begin{enumerate}
\item ({\em Single trajectory training}) For this case, the RNN is trained using a \textit{single} trajectory dataset $\{(G(i\Delta t), I( i\Delta t)\}_{i=0}^T$. The numerical advantages of employing single trajectory training stem from its relatively low computational cost throughout the optimization process. Accordingly, since the single trajectory data corresponds to a {\em specific} choice of the steaming term $F(G(t),t)$, the optimized RNN normally yields bad {\em generalization} result if we use the learned RNN to solve an IDE with different $F(G(t),t)$ term.

\item ({\em Multi-trajectory training}) In contrast with the first case, the RNN can also be trained using batch training techniques, where the input of the neural network are  \textit{multiple} trajectories generated by choosing different steaming terms $F(G(t),t)$. The training cost is obviously higher but the obtained RNN model has greater generalizability and hence can be used to predict dynamics for IDE with new steaming terms. From the operator-learning point of view, the multi-trajectory training method is preferred since the enlarged dataset contains different input-output $G(t)$ and $I(t)$, which essentially provides more test functions for learning the mapping $I: G(t) \rightarrow I(t)$.
\end{enumerate}


In accordance with these two training strategies, we also use two validation methods to detect and avoid overfitting. For the single trajectory training case, we can split the dataset $\{(G(i\Delta t), I( i\Delta t)\}_{i=0}^T$ into two parts: $\{(G(i\Delta t), I( i\Delta t)\}_{i=0}^K$ and $\{(G(i\Delta t), I( i\Delta t)\}_{i=K+1}^{T}$, and then using the first time series measure training loss used for optimization of the RNN and the second part to measure the validation error. For the multiple trajectory training, the validation error is calculated using a new dataset $\{(G(i\Delta t), I( i\Delta t)\}_{i=0}^T$ obtained by solving the IDE with streaming terms $F(G(t),t)$ that are different from those ones in the training dataset. The minimized validation error normally indicates the generalizability of the obtained RNN model. For the first case, the generalizability is reflected in the predictability of long-time dynamics. For the second case, it is also reflected in the predictability of the dynamics for IDE with new steaming terms.
\paragraph{Computational cost}  All computations are performed using the Perlmutter cluster. The login node has one AMD EPYC 7713 as the CPU and one 40GB NVIDIA A100 as the GPU. For Eqn \eqref{eq:toymodel}, a typical training with input sequence length 2000 across 750 epochs for an RNN with 2 LSTM layers and hidden size 64 would take approximately 2 hours for the multi-trajectory training and 30 minutes for the single trajectory training. Under the same setting, for Eqn \eqref{Dyson_eqn_equ}, it would take approximately 30 minutes for a multi-trajectory training and approximately 15 minutes for a single trajectory training. 
After the training, in the extrapolation phase, the computational time would be spent on the evaluation of the multi-step forward scheme and for RNN to generate new output $I(T+i\Delta t)$. Detailed runtime statistics for the numerical examples considered in this paper will be tabulated in the Results section. Generally speaking, since the RNN is made of multi-fold function compositions, the generation of the output sequence is immediate. This ensures the total computational cost in the extrapolation phase scales as of $O(T)$, in contrast with the $O(T^2)$ scaling for a normal IDE numerical solver. As a result, the entire computational cost is greatly reduced for long-time simulations. 

\section{Numerical results}\label{sec:numerical_results}
\begin{figure}[t]
    \centering
     \includegraphics[width=0.8\linewidth]{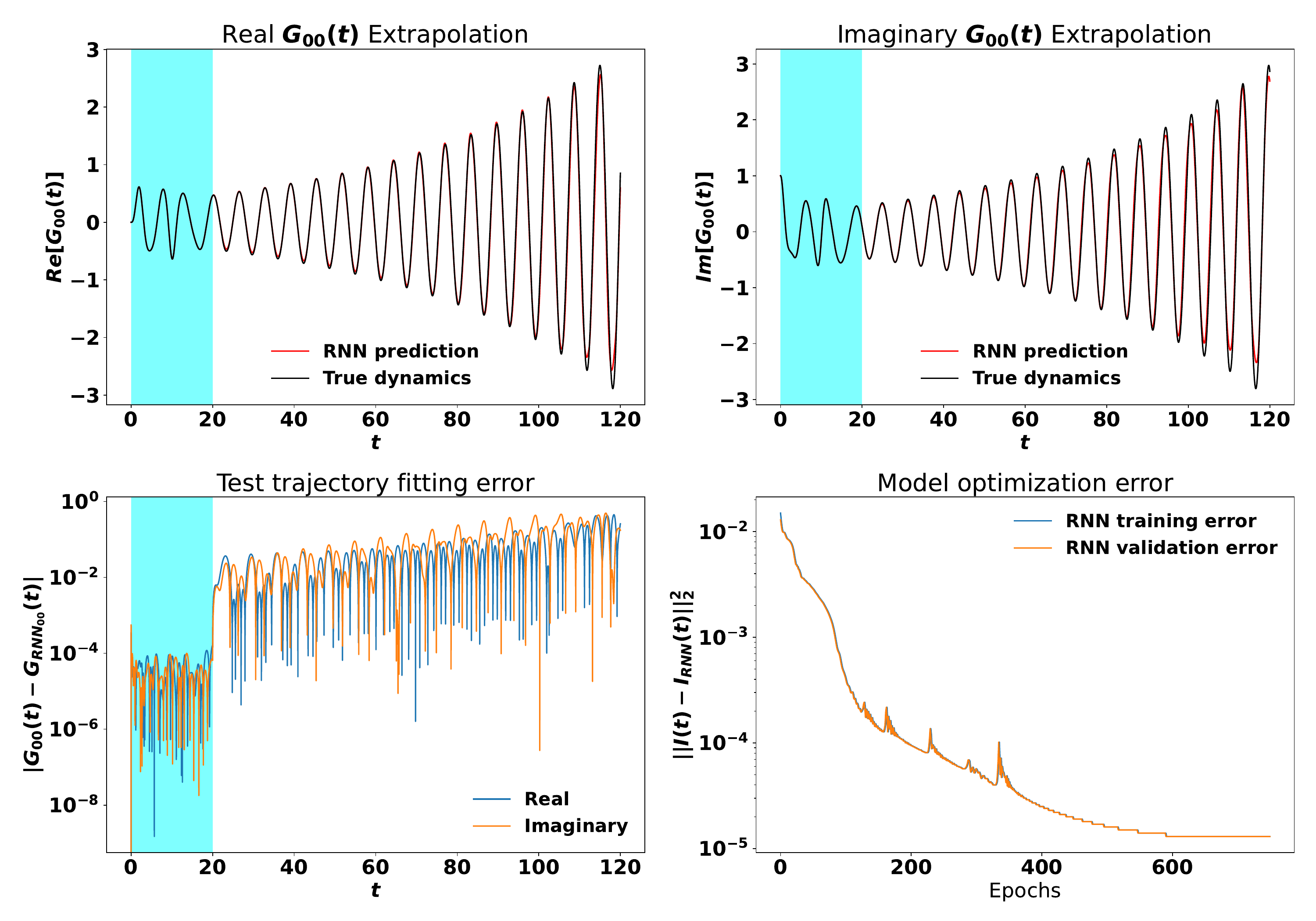}
    \caption{Single trajectory learning of IDE \eqref{eq:toymodel} where the RNN is trained on $\alpha_1 = 10,\alpha_2 = 15$, $\sigma = 2$ and $\beta =1$ and tested by extrapolating the dynamics up to $T=120$ into the future. The shaded window represents the training regime. The black curve represents the true dynamics. The red curve represents the simulated dynamics using the learned $\hat I(t)$ produced by the RNN.}
    \label{fig:single_traj_toy_model}
\end{figure}

\begin{figure*}[t]
    \centering
     \includegraphics[width=0.8\linewidth]{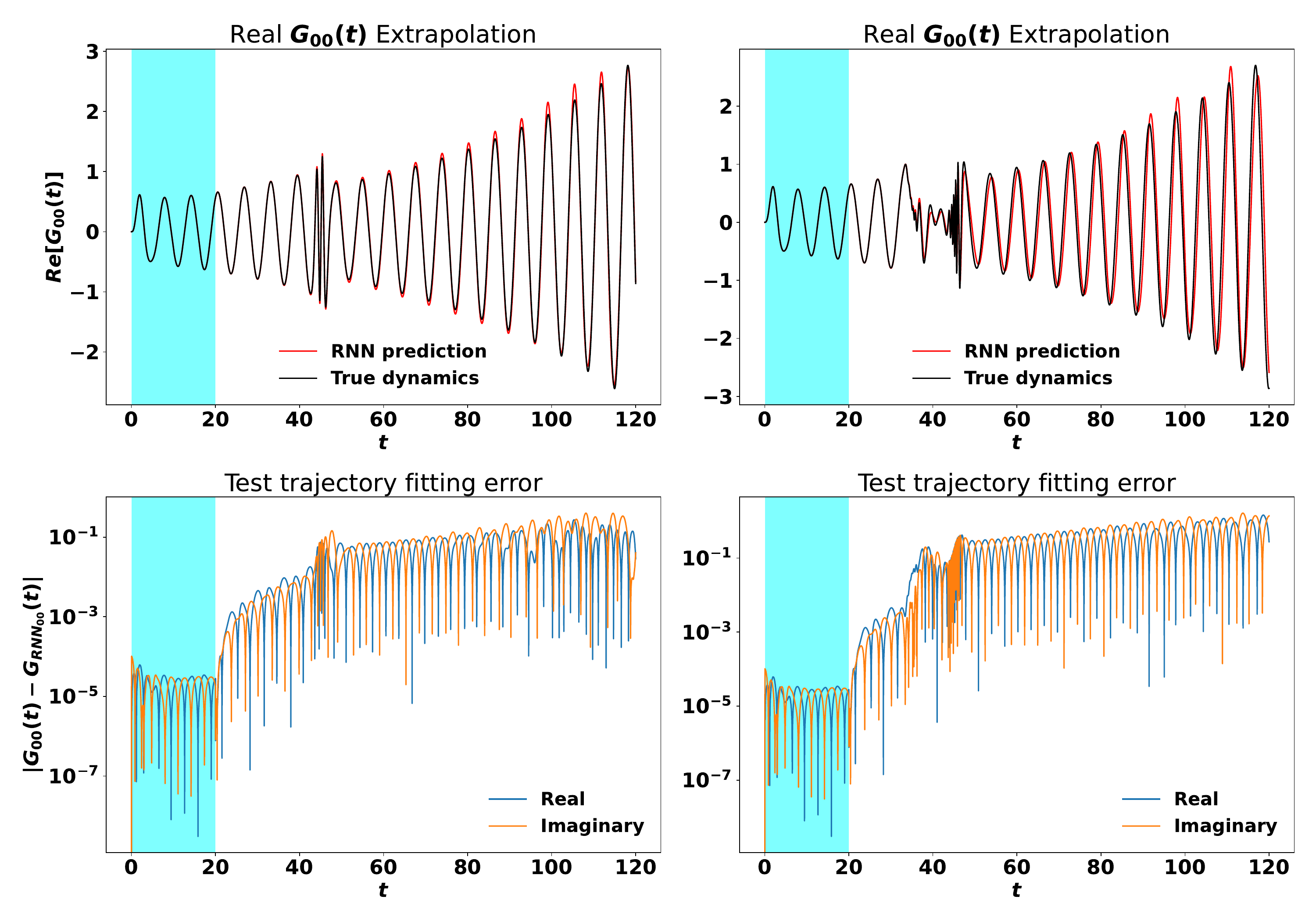}
    \caption{Multi-trajectory learning of IDE \eqref{eq:toymodel} where the RNN is trained on $\alpha_1,\alpha_2 \in [1,20]\times [1,20]$, $\sigma \in \{1,2,3,4,5\}$, $\beta=1$ and tested on $\alpha_1 = 45, \alpha_2 = 45$, $\sigma = 5$ and $\beta =1$ (\textbf{First column}) and $\alpha_1 = 45, \alpha_2 = 35$, $\sigma = 2$ and $\beta =14$ (\textbf{Second column}). The shaded window represents the training regime. The black curve represents the true dynamics. The red curve represents the simulated dynamics using the learned $\hat I(t)$ produced by the RNN model.}
    \label{fig:multi_traj_toy_model_Ct_real}
\end{figure*}

\begin{figure*}[t]
    \centering
     \includegraphics[width=0.8\linewidth]{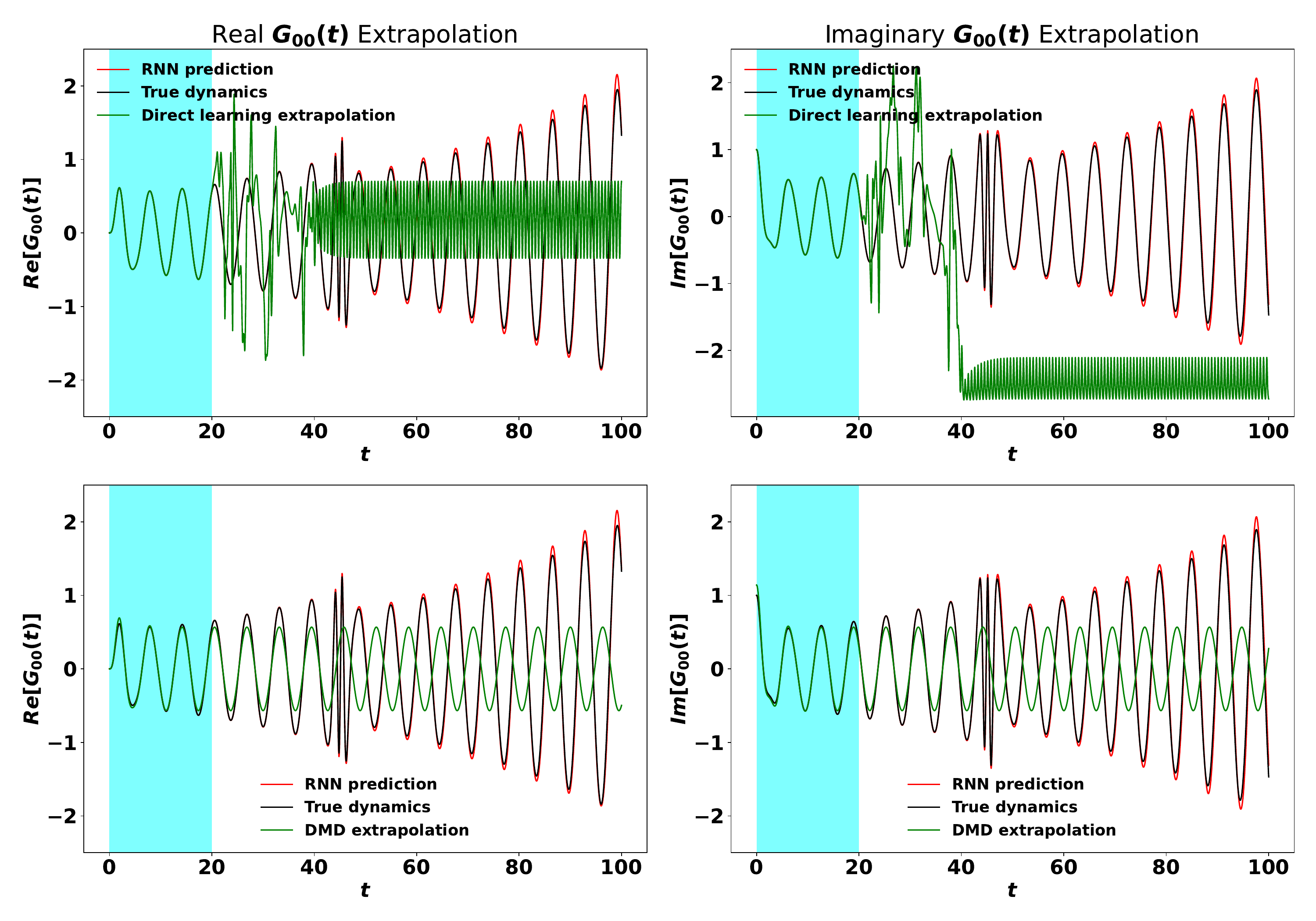}
    \caption{Multi-trajectory learning of IDE \eqref{eq:toymodel} where the RNN is trained on $\alpha_1,\alpha_2 \in [1,20]\times [1,20]$, $\sigma \in \{1,2,3,4,5\}$, $\beta=1$ and tested on $\alpha_1 = 45, \alpha_2 = 45$, $\sigma = 5$ and $\beta =1$. The shaded window represents the training regime. The black curve represents the true dynamics. The red curve represents the simulated dynamics using the learned $\hat I(t)$ produced by the RNN model. The green curve in the top row represents the simulated dynamics using the RNN model that directly predicts $G(t)$. The green curve in the bottom row represents the DMD extrapolation result.}
    \label{fig:multi_traj_toy_model_Ct_real_with_baseline_and_dmd}
\end{figure*}


To demonstrate our method, we first consider a nonlinear complex-valued IDE given by:
\begin{equation} \label{eq:toymodel}
    \frac{d}{dt}G(t) = A(t)G(t) + \int_{0}^{t}K(t-s)G(s)ds.
\end{equation} 
Here we take:
\begin{equation*}
    A(t) = -i 
            \begin{bmatrix}
                - \beta e^{\frac{-(t -\alpha_1) ^2}{\sigma}} & 1 \\
                1 & \beta e^{\frac{-(t- \alpha_2) ^2}{\sigma}}
            \end{bmatrix},
\qquad
    G(0) = 
            \begin{bmatrix}
                i & 0 \\
                0 & i 
            \end{bmatrix},
\qquad 
    K(t-s) = \cos({0.25(G(t-s)G(t-s))}).
\end{equation*}

For Eqn \eqref{eq:toymodel}, we will employ both the single trajectory training and multi-trajectory training approaches to approximate the integral operator $I(t):= \int_{0}^{t}K(t-s)G(s)ds$ and then combine the learned $I(t)$ and AB3 as the ODE solver to obtain long-time trajectories. Before we present the training details, we note in advance that the RNN training results are benchmarked in three ways:
\begin{enumerate}
\item We compare the learned and extrapolated $G(t)$ with a highly accurate numerical solution of IDE \eqref{eq:toymodel} and show the accuracy in the fitting region and RNN's predictability of the future dynamics. Due to the fact that the RNN is designed to learn the integral operator $I(t)=\int_{0}^tK(G(t-s))G(s)ds$, we also expect this predictability to be generally valid for IDE \eqref{eq:toymodel} with different $A(t)$ terms in the multi-trajectory case. 

\item We further compare our learning strategy, i.e. learning the map $I(t)=\int_{0}^tK(t-s)G(s)ds$, with two existing dynamics extrapolation methods. For the first one, we consider a direct learning strategy that uses an RNN with the same architecture while the output now changes to be the next timestep $G((i+1)\Delta t)$ value. Accordingly, we modify the loss function \eqref{eq:loss} as:
\begin{equation*}
      f(G, G_{RNN};\theta) := \frac{1}{N}\sum_{i=0}^N(G(i \Delta t) - G_{RNN}(i\Delta t))^2.
\end{equation*}
and other settings are the same. Secondly, we also compare our results with what was obtained using the dynamical mode decomposition (DMD) \cite{schmid2010dynamic,schmid2011applications,kutz2016dynamic,DMDdiag,DMDtwotime,Reeves_2023} approach. With these comparisons, one can clearly see the better generalizability of our RNN model.

\item We calculate the runtime of our simulation and compare it with what was obtained using a standard integro-differential equation solver. This would demonstrate the numerical speedup we gain using the RNN to approximate the collision integral $I(t)$. 

\end{enumerate}

\paragraph{Training details} 
For single trajectory training, we fix the modeling parameters in Eqn \eqref{eq:toymodel} to be $\alpha_1 = 10, \alpha_2 = 15, \sigma = 2,$ and $\beta = 1$. The RNN is trained by feeding in $\{G(i\Delta t)\}_{i=1}^{N}$ for $\Delta t =0.01$ and $N=2000$, and then we generate the extrapolated trajectory of $G(t)$ up to $T = 120$ ($10000$ timesteps into the future). For multi-trajectory training, the batch dataset is generated by solving IDE \eqref{eq:toymodel} with different parameters $\alpha_1,\alpha_2,\sigma,\beta$. Specifically, we prepare a dataset by choosing $\alpha_1,\alpha_2$ from the lattice grid $[1,20]\times [1,20]$, $\sigma \in \{1,2,3,4,5\}$ and $\beta=1$. This results in 2000 different trajectories.
Plotted in Figure \ref{fig:toy_model_training_data} is a reference of our input data $G(t)$ and targets $I(t)$. The displayed result is the first component of the $2$ x $2$ matrices $G(t)$ and $I(t)$. The dynamic difference between different matrix components is minimal. 

The RNN model contains 2 LSTM layers where the hidden size for each layer is 64. The input is an 8-dimensional vector that consists of the flattened $2\times 2$ matrix $G(t)$ decomposed into the real and imaginary parts. Similarly, the output is 8-dimensional, consisting of the real and imaginary parts of the $2 \times 2$ matrix collision integral $I(t)$. The Adam optimizer has an initial learning rate set to be $0.01$, with an adaptive cosine learning rate that decays accordingly with respect to total epochs used for training. This is designed to help us converge closer to the optimal solution as we progress in our optimization by taking smaller steps. The RNN is trained over 750 epochs and a batch size of 128 is chosen when we perform multi-trajectory training.

\paragraph{Results discussion} The training and testing results are summarized in Figure \ref{fig:single_traj_toy_model}-\ref{fig:multi_traj_toy_model_Ct_real_with_baseline_and_dmd}. The single trajectory training result is displayed in Figure \ref{fig:single_traj_toy_model}. We see that the RNN model is able to accurately predict the dynamics of the system using only $\frac{1}{6}$ of the total trajectory for training. In particular, the modes and amplitudes of the oscillations match well with the ground truth dynamics. The multi-trajectory training results are shown in Figures \ref{fig:multi_traj_toy_model_Ct_real}. The batch training data are obtained by varying the parameters of Eqn \eqref{Dyson_eqn_equ} to be $\alpha_1, \alpha_2 \in [1,20]\times[1,20], \sigma \in \{1,2,3,4,5\},$ and $\beta = 1$. In the first column of Figure \ref{fig:multi_traj_toy_model_Ct_real}, the learned integral operator $I(t)$ is used to solve IDE \eqref{eq:toymodel} for a new set of parameters: $\alpha_1 = 45, \alpha_2 = 45, \sigma = 5$, and $\beta = 1$, which is {\em outside} of the parameter range of the training dataset. We see that our RNN model is still able to accurately predict the dynamics of this test trajectory, despite the highly oscillatory regime of the test trajectory appearing much later, which is never seen in the dataset. This result is further highlighted in the second column of Figure 
\ref{fig:multi_traj_toy_model_Ct_real} where the test trajectory corresponds to $\alpha_1 = 45, \alpha_2 = 35, \sigma = 5$, and $\beta = 14$, and a large chirp oscillation is created after the training regime. The RNN predicting result still matches well with the true dynamics. These two test examples clearly demonstrate the generalizability of the RNN model as an integral operator.  

The result of integral operator learning is further compared with what was obtained using the direct learning approach and the DMD method. The testing results are summarized in Figure \ref{fig:multi_traj_toy_model_Ct_real_with_baseline_and_dmd}. As we introduced before, the direct learning approach used the same RNN architecture and training dataset to learn the solution map $G(i\Delta t)\rightarrow G((i+1)\Delta t)$. The DMD method can only do time extrapolation for a single trajectory therefore the training data is just the $G(t)$, $t\in[0,20]$ for fixed parameter values $\alpha_1 = 45, \alpha_2 = 45$, $\sigma = 5$ and $\beta =1$. We can see clearly from Figure \ref{fig:multi_traj_toy_model_Ct_real_with_baseline_and_dmd} that the integral operator learning strategy significantly outperforms other approaches.

Lastly, we comment on the computational cost reduction brought by the RNN integral operator learning. In Table \ref{table:rnn_cost}, we record the wall-clock runtimes used to generate different lengths of $G(t)$ in the extrapolating regime for our approach and a regular IDE solver. As we mentioned in {\em Data preparation and Loss functions}, the AB3+RNN is used to generate the extrapolated trajectory. For comparison, we employ an IDE numerical solver which consists of an Euler scheme for time integration, and Simpson's Rule for approximating the collision integral $I(t)$ (FE+SR). We see from Table \ref{table:rnn_cost} that the computational cost of querying the RNN and running AB3 is significantly lower than running FE+SR. Moreover, the overall scaling limit of our approach is of the order $O(T)$, in contrast with the $O(T^2)$ using FE+SR. We also note that for both the RNN and the FE + SR methods in Table \ref{table:rnn_cost} use $\Delta t = 0.01$.


\begin{table}
    \centering
       \begin{tabular}{ccc}
   Total simulation time & AB3 + RNN  & FE + SR  \\
\hline\hline
    20 & 0.8684s & 60.3825s  \\
    40 & 1.5028s& 166.0921s  \\
    80 & 2.5220s & 475.0468s  \\
    160 & 4.7412s & 1602.8664s  \\
\end{tabular}
\caption{Comparison of wall-clock time using the RNN method and a regular IDE solver to extrapolate dynamics with $\Delta t = 0.01$ for both methods. Here, total simulation time refers to the final time $T$ used in solving the IDE.}
\label{table:rnn_cost}
\end{table}

\begin{figure*}[t]
    \centering
     \includegraphics[width=1.0\linewidth]{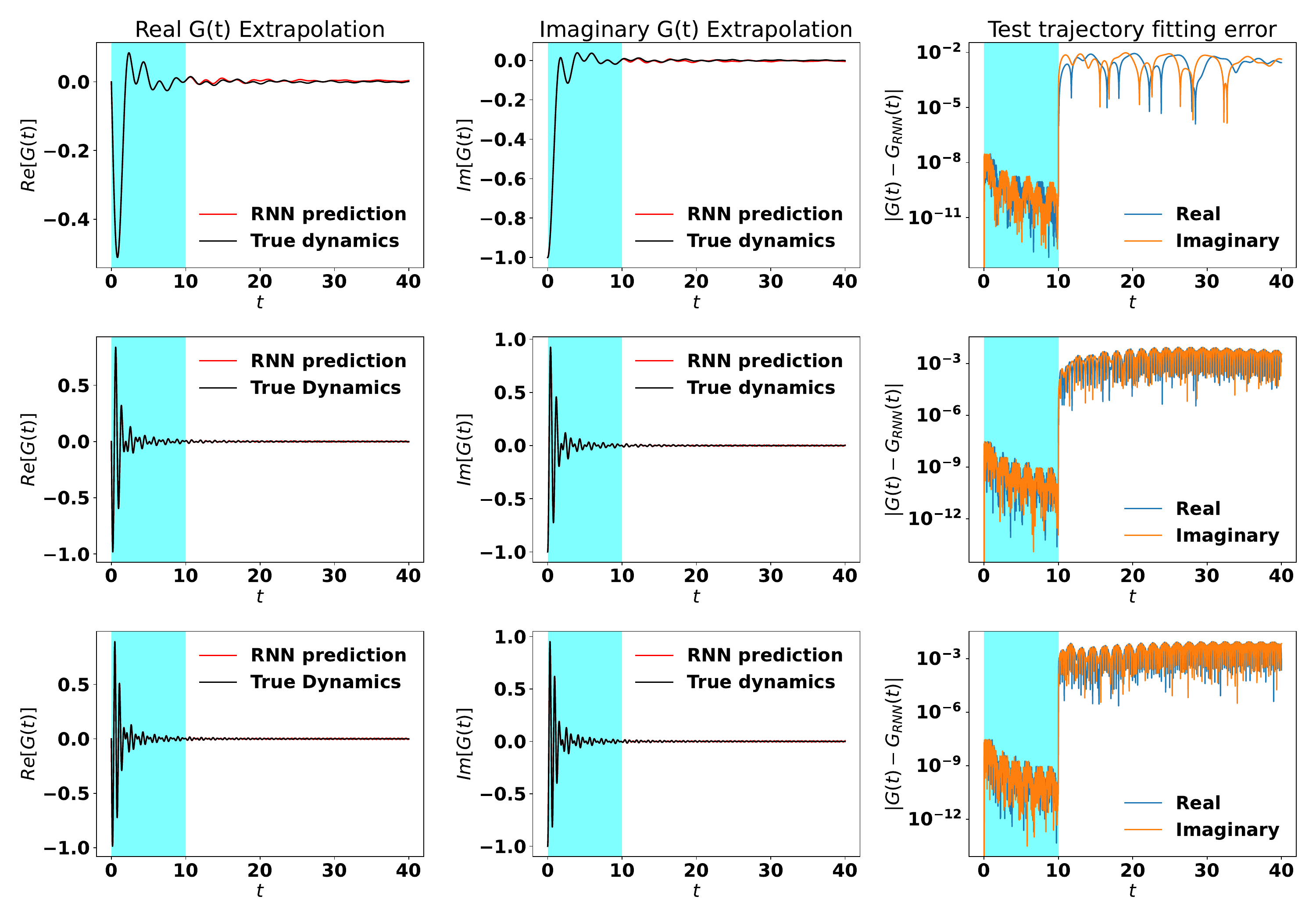}
    \caption{RNN training and extrapolating results for the Dyson's equation \eqref{Dyson_eqn_equ}. The first row displays the single trajectory training result for $h = -1,c = 1$. The second row is the multi-trajectory training results where the RNN is trained for $h \in [1,10],c = 1$, and tested on $h = 8,c = 1$. The third row uses the same multi-trajectory model, but tests on $h = 11.5, c = 1$.}
    \label{fig:total_Dyson}
\end{figure*}

\section{Application in quantum dynamics simulation}\label{sec:dyson_eqn} 
In this section, we apply the RNN model to numerically solve and extrapolate the dynamics of Dyson's equation, which is a special class of complex IDEs that is fundamentally important for the study of quantum many-body systems \cite{stefanucci2013nonequilibrium}. To benchmark our result, we consider an equilibrium Dyson's equation for hopping electrons in the Bethe lattice \cite{kaye2023fast,mahan2000many}. When $t>0$, the equation of motion reads:
\begin{align}\label{Dyson_eqn_equ}
i\partial_t G^R(t)= hG^R(t)+\int_0^tc^2G^{R}(t-s)G^R(s)ds
\end{align}
In the context of quantum many-body theory, the time-dependent quantity $G^R(t)$ is called the retarded Green's function, which contains important physical information such as the single-particle energy spectrum of the physical system. The memory kernel $K(t-s)=c^2G^R(t-s)$ is the self-energy of the system. $h,c$ are the modeling parameters that will be varied when we do multi-trajectory training. Throughout this section, the initial condition $G^{\mathrm{R}}(0) = -i$ is chosen. According to the analysis by Kaye et al.\cite{kaye2023fast}, Eqn \eqref{Dyson_eqn_equ} admits analytical solution:
\begin{align*}
G^R(t) = -ie^{-iht}\frac{J_1(2ct)}{ct},\qquad t>0
\end{align*}
where $J_1(t)$ is the Bessel function of the first kind. We will use this analytical solution to benchmark the extrapolation results generated by RNN. The training procedures are almost the same as the previous example. The slight differences are summarized in the following paragraph:
%
%
\paragraph{Training details} For IDE \eqref{Dyson_eqn_equ}, we use an RNN model with 2 layers of LSTM cells and the hidden state size for each layer is 128. The input and output are the same as it was for IDE \eqref{eq:toymodel}. For the single trajectory training case, we set $h = -1$ and $c = 1$ to build the dataset.
For the multi-trajectory training, we made a slight modification to the training procedure. We first randomly sample $h$ from the domain $[1,10]$ for 2000 times, while keeping $c = 1$. This yields a total of 2000 trajectories as the database for the subsequent RNN training. Then for each epoch, we randomly choose a batch of 10 data pairs $\{G(i\Delta t),I(i\Delta t)\}_{i=1}^{N}$ from the whole database (possibly with replacement) to optimize the RNN. In this fashion, we create a more robust model since for each epoch, the data pair $\{G(i\Delta t),I(i\Delta t)\}_{i=1}^N$ used in optimization is different.

\paragraph{Results discussion} All the training and testing results are summarized in Figure \ref{fig:total_Dyson}. In the first row, we show the single trajectory training result where the data is collected for $t\in[0,10]$ and we use the RNN to extrapolate the same trajectory up to $T = 40$ (3000 timesteps into the future). The second row shows the multi-trajectory training results where the parameters for the test trajectory are set to be $h = 8, c=1$. Note that $h$ is chosen from the sampling domain $[1,10]$ but is chosen as a parameter to be used in the training database. The third row is for the same multi-trajectory training while showing an out-of-sampling domain example where $h = 11.5, c=1$.

As we can see, both the single trajectory training and the multi-trajectory training yield precise predictions of the future-time dynamics of the Dyson's equation. Moreover, the RNN also predicts the correct asymptotic behavior of $G(t)$, i.e., $G(t)\rightarrow 0$ as $t \rightarrow \infty$. All these findings are consistent with what we found in the previous example. In comparison, we also see more accurate and robust results in the multi-trajectory case, both visually and in terms of the error plots. Since we have used many trajectories in training, the RNN is able to learn a more concrete mapping from $G(t) \rightarrow I(t)$. This leads to better predictability of the RNN on unseen dynamics, as well as greater generalizability to unseen system parameters, as shown in the final row, where $h = 11.5 \not\in [1,10]$. Nevertheless, we find in all cases that our RNN method makes accurate predictions for the dynamics well past the training regime. 

\section{Conclusion}\label{sec:conclusion}
In this paper, we introduced an RNN-based machine-learning technique that uses LSTM as the basic modeling module to learn the nonlinear integral operator in an IDE. Such a learning scheme allows us to turn an IDE to an ODE that can be solved efficiently by a standard ODE solver for a large $t$. 
We showed that a more effective way to learn a nonlinear integral operator is to include multiple training trajectories generated from different the solution of IDEs defined by different streaming terms within a small time window in the training data.  The effectiveness of this approach was demonstrated with two test examples. The generalizability of the learned operator was demonstrated by using the learned map to predict the dynamics of a new IDE that is driven by a completely different streaming term that is outside of the parameter range of the training data. Moreover, since the RNN consists of layers of function composition, it is almost immediate to generate a next timestep collision integral $I(t)$ given the input. This leads to an overall $O(T)$ scaling of computational cost (the same as an ODE solver) when we use the RNN to solve the IDE, in contrast with the $O(T^2)$ scaling of a regular IDE solver. Due to the scalability of the RNN architecture, we expect that the methodology can be generalized and used to solve high-dimensional IDEs, such as the Kadanoff-Baym equations in nonequilibrium quantum many-body theory.

\section{Acknowledgement}
This material is based upon work supported by the U.S. Department of Energy, Office of Science, Office of Advanced Scientific Computing Research and Office of Basic Energy Sciences, Scientific Discovery through Advanced Computing (SciDAC) program under Award Number DE-SC0022198.  This research used resources of the National Energy Research Scientific Computing Center, a DOE Office of Science User Facility supported by the Office of Science of the U.S. Department of Energy under Contract No. DE-AC02-05CH11231 using NERSC award BES-ERCAP0020089.





\appendix

\bibliography{sample}
\end{document}